%% file: acl_latex.tex
\pdfoutput=1

\documentclass[11pt]{article}

\usepackage[]{acl}

\usepackage{times}
\usepackage{latexsym}
\usepackage{bm}
\usepackage{multirow}
\usepackage[]{graphicx}
\usepackage{graphicx}
\usepackage{seqsplit}
\usepackage{amsmath}
\usepackage{color}
\usepackage{siunitx}
\usepackage{amsfonts}
\usepackage{empheq}
\usepackage{siunitx}

\DeclareMathOperator*{\argmin}{arg\,min}
\usepackage[T1]{fontenc}

\usepackage[utf8]{inputenc}

\usepackage{microtype}
\usepackage{chngpage}
\usepackage{float}
\makeatletter
\let\MYcaption\@makecaption
\makeatother

\usepackage{subcaption}
\makeatletter
\let\@makecaption\MYcaption
\makeatother

\title{Nearest Neighbor Non-autoregressive Text Generation}

\author{Ayana Niwa \quad Sho Takase \quad Naoaki Okazaki\\
Tokyo Institute of Technology \\
\texttt{\{ayana.niwa@nlp., sho.takase@nlp., okazaki@\} c.titech.ac.jp}
}

\begin{document}
    \maketitle
    \begin{abstract}
        Non-autoregressive (NAR) models can generate sentences with less computation than autoregressive models but sacrifice generation quality.
        Previous studies addressed this issue through iterative decoding.
        This study proposes using nearest neighbors as the initial state of an NAR decoder and editing them iteratively.
        We present a novel training strategy to learn the edit operations on neighbors to improve NAR text generation.
        Experimental results show that the proposed method (NeighborEdit) achieves higher translation quality (1.69 points higher than the vanilla Transformer) with fewer decoding iterations (one-eighteenth fewer iterations) on the JRC-Acquis En-De dataset, the common benchmark dataset for machine translation using nearest neighbors. We also confirm the effectiveness of the proposed method on a data-to-text task (WikiBio).
        In addition, the proposed method outperforms an NAR baseline on the WMT'14 En-De dataset. We also report analysis on neighbor examples used in the proposed method.
\end{abstract}
    \input{sections/1-introduction}

    \input{sections/2-3-method}
    \input{sections/4-exp}

    \input{sections/5-related_works}

    \input{sections/6-conclusion}

    \bibliography{anthology,custom}
    \bibliographystyle{acl_natbib}
    \clearpage
    \appendix

    \section{Appendix\label{sec:appendix}}
    \begin{table}[th]
        \centering
        \small
        \begin{tabular}{lrrr}\hline
         Dataset    &  Train & Validation & Evaluation 
         \\\hline
        JRC-Acquis& 475,330& 2,945& 2,937
        \\
        WikiBio &582,659 &72,831&72,831
        \\
        WMT'14 &3,961,179&3,000&3,003
        \\\hline
        \end{tabular}
        \caption{Dataset statistics\label{tab:data_stats}}
    \end{table}
    \subsection{Dataset Details}
    Table~\ref{tab:data_stats} lists the statistics of the datasets in the experiments.
     We did not use the distilled dataset~\cite{kim-rush-2016-sequence} in this study.

    \subsection{Model Details and Hyperparameters\label{sub:model_settings}}
    For Neighboredit, we set $d_\mathrm{model}=512$, $d_\mathrm{hidden}=2048$, $n_\mathrm{head}=8$, and $n_\mathrm{layer}=6$.
    In the proposed oracle policy in Section~\ref{subsec:oracle_policy}, we set the hyperparameter $\alpha$ to $0.6$ for all settings.
    In addition, we set $\beta$ to $0.3$ for all retrieval methods in JRC-Acquis, TableMatch in WikiBio, and TFIDF in the WMT'14. For SentVec+TFIDF in the WMT'14, we set to $1.0$.
    These are the values automatically determined by the development data as well as other hyperparameters.
    A threshold value of 1.0 means that special tokens are used for initialization instead of neighbors for all data during inference. However, during training, Neighboredit uses the neighbors to learn the deletion operation, which led to improved performance in Table~\ref{tab:wmt_main_result}.
    For the policy classifier, we set $K_\mathrm{max}$ to 255.
    Embeddings of the source and target sides are shared. 
    Furthermore, the parameters of the three edit operations in the decoder are shared.

   \subsection{Neighbor Retrieval Details}\label{subsec:retrieval_details}
    For TFIDF, we first created high-dimensional sparse TFIDF vectors and reduced the dimension to 512 by using singular value decomposition (SVD).
    After the neighbors are retrieved by faiss, we reranked the Top-50 neighbor candidates by calculating the exact cosine similarity of the uncompressed TFIDF vectors between the candidates and source sentences to reduce information loss for the TFIDF.
    For SentVec, we use the vectors (512 dimensions) computed by the pre-trained encoder as they are.
    Nearest neighbor retrieval was much faster than the decoding process of the NAR model; thus, the decoding overhead by nearest-neighbor retrieval is negligible.

    \subsection{Training Details}

    We trained BPE~\cite{sennrich-etal-2016-neural} to construct a vocabulary with 20,000 joint operations for the JRC-Acquis and 40,000 joint operations for the WMT'14.
    For the WikiBio, we did not use the subword tokenization.

    We followed the weight initialization scheme of BERT~\cite{devlin-etal-2019-bert}.
    We used Adam~\cite{DBLP:journals/corr/KingmaB14} optimizer with $\beta=(0.9, 0.98)$.
    For regulation, we set the dropout rate of 0.3 and the weight decay of 0.01.
    We linearly increased the learning rate from $1\times 10^{-7}$ to $5\times10^{-4}$ in the initial 10,000 steps and then used the learning rate decay of square root.
    All models were trained on 4 NVIDIA Tesla P100 GPUs for the JRC-Acquis and WikiBio, and 16 NVIDIA V100 GPUs for the WMT'14.
 
    \subsection{Evaluation Details}
    We selected the checkpoints according to the BLEU scores of the development set.
    SacreBLEU hash is \seqsplit{BLEU+case.mixed+numrefs.1+smooth.exp+tok.13a+version.1.5.1}.
    For a statistical significance test, we adopted a paired bootstrap resampling~\cite{dror-etal-2018-hitchhikers}.

    \subsection{Output examples\label{subsec:output_ex}}
    Figure \ref{tab:output_example} shows output examples of Levenshtein Transformer and NeighborEdit.
    The proposed method could generate correct sentences with fewer iterations by removing unnecessary words from the neighbor examples and filling the placeholders.
    \begin{figure*}[ht]
        \centering
        \small
        \begin{tabular}{ll}
            \hline
            Source                       & ( 48 ) Capacity utilisation increased by 56 \% between 1996 and the IP.                                                                                                                         \\
            Target                       & ( 48 ) Die Kapazitätsauslastung stieg von 1996 bis zum UZ um 56 \%.                                                                                                                             \\\hline
            \multicolumn{2}{c}{\textbf{Levenshtein Transformer}} \\\hline
            Deletion 1                   &                                                                                                                                                                                                 \\
            \multirow{2}{*}{Insertion 1} & \fontsize{8pt}{0}\selectfont
            \textcolor{red}{\texttt{[PLH][PLH][PLH][PLH][PLH][PLH][PLH][PLH][PLH][PLH][PLH][PLH][PLH][PLH][PLH][PLH]}} \\
            & \textcolor{red}{( 48 ) Die Kapazitätsauslastung stieg von 1996 bis dem \textcolor{blue}{UZ} UZ um 56 \%.}                                                                                       \\
            Deletion 2                   & ( 48 ) Die Kapazitätsauslastung stieg von 1996 bis dem UZ um 56 \%.                                                                                                                             \\
            \multirow{2}{*}{Insertion 2} & ( 48 ) Die Kapazitätsauslastung stieg von 1996 bis \textcolor{red}{\texttt{[PLH]}} dem UZ um 56 \%.                                                                                             \\
            & ( 48 ) Die Kapazitätsauslastung stieg von 1996 bis \textcolor{red}{zu} dem UZ um 56 \%.                                                                                                         \\\hline
            \multicolumn{2}{c}{\textbf{NeighborEdit}} \\\hline
            Neighbor (En)                & ( 56 ) Export prices of the two cooperating Thai producers have increased by 6 \% between 1996 and the IP.                                                                                      \\
            Neighbor (De) &
            ( \textcolor{blue}{56} )
            Die \textcolor{blue}{Ausfuhrpreise der beiden kooperierenden thailänd@@ ischen Hersteller stiegen} \\
            & \hspace{3cm} von 1996 bis zum \textcolor{blue}{Untersuchungszeitraum} um \textcolor{blue}{6} \%                                                                                                 \\
            Deletion 1                   & ( ) Die von 1996 bis zum um \%.                                                                                                                                                                 \\
            \multirow{2}{*}{Insertion 1} & ( \textcolor{red}{\texttt{[PLH]}} ) Die \textcolor{red}{\texttt{[PLH]}} \textcolor{red}{\texttt{[PLH]}} von 1996 bis zum \textcolor{red}{\texttt{[PLH]}} um \textcolor{red}{\texttt{[PLH]}} \%. \\
            & ( \textcolor{red}{48} ) Die \textcolor{red}{Kapazitätsauslastung stieg} von \textcolor{red}{1996} bis zum UZ um \textcolor{red}{56} \%.                                                         \\\hline
        \end{tabular}
        \caption{An example of JRC-Acquis En-De translation by Levenshtein Transformer and NeighborEdit (proposed). We present inserted tokens in \textcolor{red}{red color} and deleted tokens in \textcolor{blue}{blue color}. The proposed method could generate the correct sentence with a single iteration. \label{tab:output_example}}
    \end{figure*}

\end{document}

%% file: sections/1-introduction.tex
    \section{Introduction}
    Non-autoregressive (NAR) models have gained popularity lately~\cite{gu2018nonautoregressive,ghazvininejad-etal-2019-mask,lee-etal-2020-iterative,qian-etal-2021-glancing}.
    Autoregressive (AR) generation models ~\cite{Sutskever:NIPS2014} need to iterate computations of forward propagation from the input to output layers of the decoder multiple times for all generated tokens, whereas NAR models predict multiple tokens simultaneously.
    Hence, NAR decoding is faster as only one forward propagation computation is required.
    However, due to the inherent difficulty in capturing the dependencies between generated tokens, the generation quality of NAR models is worse than that of AR models.

    The primary solution to this problem is to repeat decoding processes including edit operations, e.g., token substitution~\cite{ghazvininejad-etal-2019-mask,qian-etal-2021-glancing}, insertion~\cite{stern2019insertion}, and deletion/insertion~\cite{NEURIPS2019_675f9820}.
    Repeating generation processes decreases the dependencies between output tokens by learning the conditional distribution over the generated tokens~\cite{gu-kong-2021-fully}.
    However, during the inference, the decoder must start the generation from scratch (or special tokens), which leads to a low-quality sentence generated in the first iteration.
    Moreover, \citet{huang2022improving} have demonstrated that if the initial sentence generated is of low quality, it is difficult to recover even after several iterations.
    Therefore, it is preferable to begin the generation process with a high-quality sentence to improve performance.

    Therefore, this paper proposes using the \emph{nearest neighbor} as the initial state of the NAR decoder. 
    The proposed method, \textbf{NeighborEdit}, retrieves the nearest neighbor of an input sentence and edits it to generate the output sentence.    
    Thus, NeighborEdit can start the generation with a sentence that is close to that of the output.
    As generating from a neighbor example is easier than generating from scratch, we anticipate an improved generation quality with fewer iterations.
    Consider the following example where the English sentence ``I have an apple.'' is translated into the German sentence ``Ich habe einen Apfel.'' 
    NeighborEdit first retrieves the nearest neighbor of the English sentence from the training data, ``I have a banana. -- Ich habe eine Banane.''
    The method then starts with the retrieved German sentence, deletes the words `eine' and `Banane' and inserts the words `einen' and `Apfel,' to complete the translation.
    
    Traditionally, nearest neighbors have been utilized in various tasks, including part-of-speech tagging~\cite{daelemans-etal-1996-mbt} and example-based machine translation~\cite{Nagao:84}.
    Recently, nearest neighbors have been successfully applied to AR models, particularly to neural machine translation~\cite{gu2018search,xu-etal-2020-boosting, khandelwal2021nearest}, text summarization~\cite{cao-etal-2018-retrieve, peng-etal-2019-text}, and data-to-text generation~\cite{wiseman-etal-2021-data}.
    We adapt the pioneering idea to refine and improve NAR models.
    
    NeighborEdit is regarded as one of the edit-based NAR models that have been actively studied in recent years.
    In reality, it is not always possible to retrieve a neighbor close to the output sentence, and learning and applying editing operations to a distant sentence can be challenging.
    Therefore, we also propose a training strategy based on the lexical difference between the neighbor and the output sentences by mixing two policies, neighbor- and target-centric policies (described in Section~\ref{subsec:oracle_policy}).
    For the base architecture, we adopt the Levenshtein transformer~\cite{NEURIPS2019_675f9820}, which permits two edit operations on input sentences: token deletion and token insertion. 
    This model does not require that the output length be fixed, whereas most NAR models require that the output length be predicted.    %
    
    Experimental results show that the proposed method (NeighborEdit) achieves higher translation quality (1.69 points higher than the vanilla Transformer) with fewer decoding iterations (one-eighteenth fewer iterations) on the JRC-Acquis En-De dataset, the common benchmark dataset for machine translation using nearest neighbors. We also confirm the effectiveness of the proposed method on a data-to-text task (WikiBio).
    In addition, the proposed method outperforms an NAR baseline on the WMT'14 En-De dataset. Lastly, we report analysis on neighbor examples utilized by the proposed method.

%% file: sections/2-3-method.tex
    \section{Problem Formulation and Preliminary}
    \begin{figure*}[t]
        \centering
        \includegraphics[width=\linewidth]{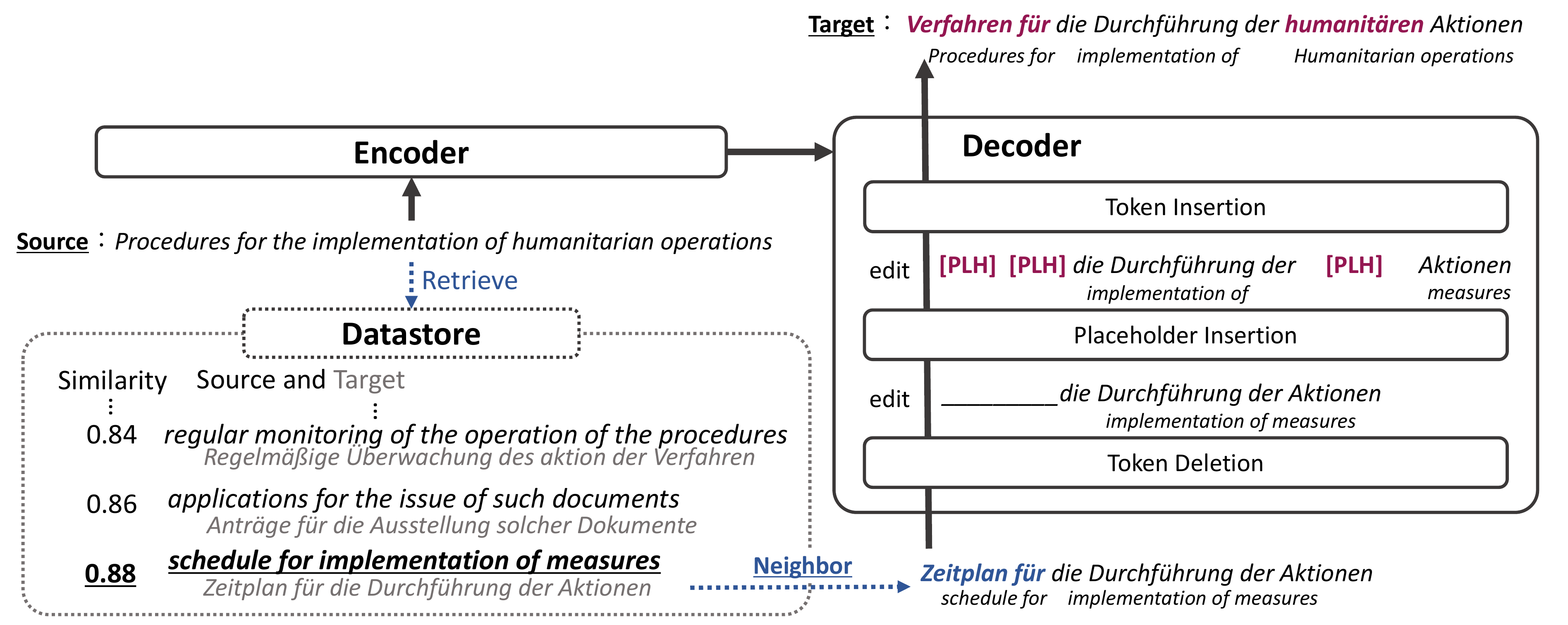}
        \caption{An illustration of our proposed method in the machine translation (MT) task. When translating the English sentence, ``Procedures for the implementation of humanitarian operations'' to the German sentence, ``Verfahren für die Durchführung der humanitären Aktionen,'' we set the neighbor ``Zeitplan für die Durchführung der Aktionen
            (Schedule for implementation of measures)'' as the initial state of the decoder and repeat the edit operations.}
        \label{fig:model_outline}
    \end{figure*}
    We consider a translation task for simplicity.
    A target sentence $\bm{y}^*$ is generated from a source sentence $\bm{x}$ and a neighbor example $\bm{z}_0$.
    Here, we find the sentence $\bm{x}'$ that is closest to the source sentence $\bm{x}$ in the datastore, and let $\bm{z}_0$ be the target sentence paired with $\bm{x}'$.
    
    \subsection{Problem Formulation}\label{subsubsec:problem_form}
    We train the model using imitation learning, a type of reinforcement learning, to mimic the sequence of actions of an expert.
    We cast the sequence generation task as a Markov Decision Process defined by the tuple $(\mathcal{Y}, \mathcal{A}, \mathcal{E}, \mathcal{R}, \bm{z}_0)$. Here, $\bm{y}\in\mathcal{Y}$ is a partially constructed sequence during the decoding process, which we refer to as a \textit{canvas} \cite{DBLP:conf/icml/SternCKU19,wiseman-etal-2021-data}.
    $\mathcal{A}$ is a set of actions (edit operations), $\mathcal{E}$ is the environment that receives a canvas $\bm{y}$ and action $\bm{a}$ and yields an edited sequence, and $\mathcal{R}$ is a reward function.

    At the $k$-th iteration of decoding, the model receives canvas $\bm{y}^{k-1}$ (of $n+1$ tokens)\footnote{$y_0 = \mbox{\tt <s>}$ and $y_n = \mbox{\tt </s>}$, which represent the beginning and end of a sequence. The value $n$ may change during the iteration process via token deletion and insertion.} edited at the previous iteration.
    The model then selects the action $\bm{a}^k$ to generate a new canvas $\bm{y}^k=\mathcal{E}(\bm{y}^{k-1}, \bm{a}^k)$ (of $n+1$ tokens) and obtains the reward $r^k=\mathcal{R}(\bm{y}^k)$.
    The policy $\pi$ to select the action $\bm{a}^k$ is defined as the mapping from the decoder output $(\bm{h}_0, \bm{h}_1, ..., \bm{h}_n)$ for the input canvas $\bm{y}^{k-1}$ into action space $\mathcal{A}$.
    Generally, the policy $\pi$ is approximated by linear classifiers with parameter $\theta$.

    \subsection{Recap: Levenshtein Transformer}
     NeighborEdit adopts Levenshtein Transformer~\cite{NEURIPS2019_675f9820}, an edit-based non-autoregressive encoder-decoder model, as the base architecture.
    More specifically, policy $\pi$ consists of three parts: (1) token deletion classifier $\pi_{\theta}^\mathrm{del}$, (2) placeholder (\texttt{[PLH]}) insertion classifier $\pi_{\theta}^\mathrm{plh}$, and (3) token classifier $\pi_{\theta}^\mathrm{tok}$.
    
    \subsubsection{Policy Classifier \label{subsubsec:policy_cls}}

    \textbf{(1) Deletion classifier} $\pi_{\theta}^\mathrm{del}$ predicts whether each token $y_i$ ($i \in \{1, \dots, n-1\}$) should be deleted ($d_i=0$) or kept ($d_i=1$) as a binary classification.
    \begin{gather}
        \pi_{\theta}^\mathrm{del} (d_i|i,\bm{y}) =  \mathrm{softmax}(\bm{h}_i  \bm{A})\\\label{eq:del}
        \bm{A} \in
        \mathbb{R}^{d_{\mathrm{model}} \times 2},\; \forall i: i \in \{1, .., n-1\} \notag
    \end{gather}

    \textbf{(2) Placeholder insertion classifier} $\pi_{\theta}^\mathrm{plh}$ predicts the number of placeholders (\texttt{[PLH]}) to be inserted $p_i \in \{0, \dots, K_\mathrm{max}\}$ at every consecutive pair of positions.
    When $p_i=0$, the model inserts no placeholder between $y_{i}$ and $y_{i+1}$.
    \begin{gather}
        \pi_{\theta}^\mathrm{plh} (p_i|i,\bm{y}) =  \mathrm{softmax}([\bm{h}_{i}; \bm{h}_{i+1}])\bm{B})\\
        \bm{B}\in\mathbb{R}^{(2d_{\mathrm{model}})\times(K_{\mathrm{max}}+1)},\; \forall i: i \in \{0, .., n-1\}\notag
    \end{gather}
    Here, $[\bm{a};\bm{b}]$ represents a concatenation of the vectors ${\bm{a}}$ and ${\bm{b}}$.

    \textbf{(3) Token classifier} $\pi_{\theta}^\mathrm{tok}$ predicts a token $t_i$ for filling the placeholder \texttt{[PLH]}.
    \begin{gather}
        \pi_{\theta}^\mathrm{tok} (t|i,\bm{y}) =  \mathrm{softmax}(\bm{h}_i \bm{C}),\label{eq:tok_ins}\\
        \bm{C}\in\mathbb{R}^{d_{\mathrm{model}}\times |\mathcal{V}|}\; \forall i: y_i=\texttt{[PLH]} \notag
    \end{gather}
    
    In total, the policy classifier has the parameter $\theta = (\bm{A}, \bm{B}, \bm{C})$.
    The model receives the canvas $\bm{y}$ and performs three edit operations in the following order: (1) token deletion $d_i\; (i \in \{0, ..., n\})$, (2) placeholder insertion $p_i\; (i \in \{0, ..., n-1\})$, and (3) token insertion $t_i\; (\forall i: y_i=\texttt{[PLH]})$.
    
     \subsubsection{Training}\label{subsubsec:training}
    We train the model parameters by minimizing the sum of deletion $\mathcal{L}_{\theta}^{\mathrm{del}}$ and insertion $ \mathcal{L}_{\theta}^{\mathrm{ins}}$ losses.
    \begin{align}
        \label{eq:loss_function}
        \mathcal{L}=& \mathcal{L}_{\theta}^{\mathrm{del}}  + \mathcal{L}_{\theta}^{\mathrm{ins}}\\
        \mathcal{L}_{\theta}^{\mathrm{del}}=& -\sum_{y^i\in \bm{y}_{\mathrm{del}},\; d_i^*\sim \bm{d}^*}\mathrm{log}\pi_{\theta}^{\mathrm{del}}(d_i^*|i, \bm{y}_{\mathrm{del}})\\
        \mathcal{L}_{\theta}^{\mathrm{ins}}=& -\sum_{y^i\in \bm{y}_{\mathrm{ins}},\; p^*_i\in \bm{p}^*} \mathrm{log} \pi_{\theta}^{\mathrm{plh}}(p_i^*|i, \bm{y}_{\mathrm{ins}}) \notag \\
        & -\sum_{t^*_i\in \bm{t}^*}\mathrm{log}\pi_{\theta}^{\mathrm{tok}}(t_i^*|i, \bm{y}'_{\mathrm{ins}})
    \end{align}
    Here, $\bm{y}_{\mathrm{ins}}$ and $\bm{y}_{\mathrm{del}}$ are the canvases of the insertion target and deletion target, respectively. Also, $\bm{y}'_{\mathrm{ins}}$ represents the canvas after inserting \texttt{[PLH]} to $\bm{y}_{\mathrm{ins}}$.
    The expert operations $\bm{a}^*\in \{\bm{d}^*, \bm{p}^*, \bm{t}^*\}$ are chosen to minimize the Levenshtein distance $\mathcal{D}$~\cite{levenshtein1966binary} between expert $\bm{e}$ and edited canvas $\mathcal{E}(\bm{y}, \bm{a})$.
    \begin{align}
        \bm{d}^* &= \argmin_{\bm{d}}\mathcal{D}(\bm{e}, \mathcal{E}(\bm{y}_{\mathrm{del}}, \bm{d}))\\
        \bm{p}^*, \bm{t}^* &= \argmin_{\bm{p}, \bm{t}}\mathcal{D}(\bm{e}, \mathcal{E}(\bm{y}_{\mathrm{ins}}, \{\bm{p}, \bm{t}\})
    \end{align}
    Please refer to \citet{NEURIPS2019_675f9820} for more details of Levenshtein Transformer.
    \section{Proposed Method\label{sec:proposed_method}}
    As shown in Figure~\ref{fig:model_outline}, the proposed method first retrieves the nearest neighbors from the datastore consisting of all parallel sentences in the training set (Section~\ref{subsec:retrieval}).
    The method then generates a sentence by deleting and inserting tokens on the retrieved neighbor (Section~\ref{subsec:edit_model}).

    \subsection{Edit Operations on Neighbors\label{subsec:edit_model}}
    NeighborEdit is an NAR model based on Transformer~\cite{vaswani2017attention}.
    Given the source sentence $\bm{x}$ as the input and a retrieved neighbor $\bm{z}_0$ as \emph{the initial state of the decoder}, the proposed model edits the neighbor by deleting and inserting tokens repeatedly.
    We focus on incorporating the retrieved neighbors $\bm{z}_0$ into the decoding process.

    \subsubsection{Oracle Policy for NeighborEdit\label{subsec:oracle_policy}}
    During imitation learning, the model learns the parameterized policy $\pi_{\theta}$ to mimic the actions of an oracle policy $\pi^*$.
    We designed a novel strategy to obtain an oracle policy to edit the neighbor $\bm{z}_0$ to a target sentence effectively.
    
    The basic idea of the policy is to reproduce the edit operations on the intermediate sequence that appears in the iteration during inference (the partially edited neighbor).
    Therefore, our policy focuses on the lexical difference between a neighbor and target sentence.
    However, if their similarity is too low, it is necessary to rewrite a sentence into a completely different sentence.
    The policy of complete rewriting is too difficult to learn.
    To alleviate this issue, we design a policy by mixing the \textbf{neighbor}-centric policy $\pi^*_\mathrm{n}$ and  \textbf{target}-centric policy $\pi^*_\mathrm{t}$.
    The former utilizes the lexical similarity and difference between the neighbor $\bm{z}_0$ and the target sentence $\bm{y}^*$, whereas the latter uses the target sentence $\bm{y}^*$ only.
     During training, the model randomly chooses  $\pi^*_\mathrm{n}$ or $\pi^*_\mathrm{t}$ for each batch.
    Furthermore, when the similarity between the neighbor and target sentence $\mathrm{sim}(\bm{z}_0, \bm{y}^*)$ is less than a threshold $\beta$, the policy $\pi^*_\mathrm{n}$ for the insertion switches to the $\pi^*_\mathrm{t}$ for each instance\footnote{We do not set a threshold for the deletion in the $\pi^*_\mathrm{n}$ because the model is sure to remove unnecessary tokens from neighbors during inference.}.
    In summary, our policy $\pi^*$ mixies the policies  $\pi^*_\mathrm{n}$ and  $\pi^*_\mathrm{t}$ as follows:
    \begin{align}
        \pi^* &= (\pi^* \text{for deletion}, \pi^* \text{for insertion})\\\notag
        &=    \left\{
    \begin{array}{ll}
    (\pi^*_{\mathrm{n}},  \pi^*_{\mathrm{n}})& (u<\alpha\wedge \mathrm{sim}(\bm{z}_0, \bm{y}^*)>\beta)\\
    (\pi^*_{\mathrm{n}},  \pi^*_{\mathrm{t}})&(u<\alpha \wedge \mathrm{sim}(\bm{z}_0, \bm{y}^*)\leq\beta)\\
    (\pi^*_{\mathrm{t}},  \pi^*_{\mathrm{t}}) & (\text{otherwise}),
    \end{array}
    \right. \label{eq:switch}
    \end{align}
     where $\alpha \in [0,1]$ is a hyperparameter and $u$ is a value sampled from the uniform distribution of $[0,1]$.
     Here, we define  $\mathrm{sim}(\bm{z}_0, \bm{y}^*)$ as the token overlap ratio, the number of overlapped tokens divided by the number of tokens in the neighbor.
    
    By the above training strategy, our policy enables the model to delete unfavorable tokens from the nearest neighbor and insert necessary tokens.
    Given a canvas $\bm{y}$, we define the oracle policy $\pi^*$ to generate expert $\bm{e}$ as follows.

    \paragraph{Deletion: $\pi^*(\bm{y}_{\mathrm{del}})$}
    This edit operation removes unnecessary tokens from the canvas $\bm{y}_{\mathrm{del}}$.
    It is essential to delete all unnecessary tokens from the retrieved neighbor $\bm{z}_0$.
    Furthermore, unnecessary tokens that the model inserted in the previous iteration need to be removed.
    Hence, we train the model to delete tokens from $\bm{z}_0$ that do not appear in the target sentence $\bm{y}^*$ or unnecessarily inserted tokens from $\mathcal{E}(\bm{y}^{'}, (\tilde{p},\tilde{t}))$.
    \begin{gather}
        \pi^*_{\mathrm{n}}(\bm{z}_0) =  \mathrm{M}(\bm{y}^{*}, \bm{z}_0) \\
        \pi^*_{\mathrm{t}}( \mathcal{E}(\bm{y}^{'}, (\tilde{p},\tilde{t}))) =  \mathrm{M}(\bm{y}^{*}, \bm{y}_\mathrm{del}) \\
        \quad\tilde{p}\sim\pi_{\theta}^{\mathrm{plh}}, \tilde{t}\sim\pi_{\theta}^{\mathrm{tok}}\notag
    \end{gather}
    The function $\mathrm{M}(\bm{a},\bm{b})$ yields the common token subsequence between $\bm{a}$ and $\bm{b}$.
    For example, $\mathrm{M}$(``ABCDE,'' ``ARE'') returns ``AE.''

    \paragraph{Insertion: $\pi^*(\bm{y}_{\mathrm{ins}})$}
    This operation inserts necessary tokens into the canvas $\bm{y}_\mathrm{ins}$ to generate $\bm{y}^*$.
   
    \begin{gather}
        \pi^*_{\mathrm{n}}(\mathrm{M}(\bm{y}^*, \mathcal{E}(\bm{z}_0, \tilde{d}))) =  \bm{y}^* \;
        \tilde{d}\sim\pi_{\theta}^{\mathrm{del}}\\
        \pi^*_{\mathrm{t}}(  \mathcal{E}(\bm{y}^*, \tilde{d})) =  \bm{y}^* \;
        \tilde{d}\sim\pi^{\mathrm{RND}} 
    \end{gather}
    Here, $\pi^{\mathrm{RND}}$ is a policy of random token deletion.
    $\mathcal{E}(\bm{y}^*, \tilde{d})$ presents a target sequence with some tokens deleted at random (based on $\pi^{\mathrm{RND}}$).
    The intention of $\mathrm{M}(\bm{y}^*, \mathcal{E}(\bm{z}_0, \tilde{d}))$ is to delete tokens in the nearest neighbor based on the policy of $\pi_{\theta}^{\mathrm{del}}$ and insert necessary tokens that appear in $\bm{y}^*$ but not in the deleted sequence.

    \subsubsection{Inference\label{subsubsec:inference}}
    The model receives the initial state (neighbors) and generates a sentence by repeatedly selecting and applying the most probable edit operation to the canvas $\bm{y}$.
    We terminate the process when we reach the maximum number of iterations or when the output sequences from two consecutive iterations are identical.
    
    However, the initial state may hurt the performance if no close example exists in the datastore.
    Therefore, this study employs a switching approach as well as the oracle policy. 
    Specifically, if the similarity of the neighbor on the source side $\mathrm{sim}(\bm{z}_0, \bm{x})$ is less than the threshold $\beta$ (used in Equation~\eqref{eq:switch}), we do not initialize the decoder with the neighbors but do with special tokens \texttt{<s></s>} instead, similarly to common NAR models.

    \subsection{Neighbor Retrieval\label{subsec:retrieval}}
    To obtain the nearest neighbor $\bm{z}_0$, we retrieve the most similar example to a given input sentence from the datastore.
    The datastore contains all the parallel sentences in the training data.
    Existing retrieval methods for nearest neighbors can be divided into two major types: lexical matching~\cite{xu-etal-2020-boosting, gu2018search, bulte-tezcan-2019-neural} and distributional similarity~\cite{khandelwal2021nearest, borgeaud2021improving}.

    \paragraph{Lexical matching}
    Previous studies explored fuzzy matching and n-gram matching~\cite{xu-etal-2020-boosting, gu2018search, bulte-tezcan-2019-neural}.
    This study uses token frequencies, as in \citet{peng-etal-2019-text}, because this method scales well with large amounts of data.
    Specifically, we define a similarity score $S_{\mathrm{TFIDF}}$ for sentences $s_i$ and $ s_j$ using the cosine similarity of the TFIDF vector $\text{tfidf}(\cdot)$.
    \begin{align}
        S_{\mathrm{TFIDF}}(s_i, s_j)= \frac{\text{tfidf}(s_i)\cdot \text{tfidf}(s_j)}{\|\text{tfidf}(s_i)\| \|\text{tfidf}(s_j)\|}
    \end{align}

    \paragraph{Distributional similarity}
    We use the cosine similarity $S_{\mathrm{SentVec}}$ of the sentence vector extracted from an off-the-shelf encoder.
    \begin{align}
        S_{\mathrm{SentVec}}(s_i, s_j) = \frac{\bm{h_i}\cdot \bm{h_j}}{\|\bm{h_i}\| \|\bm{h_j}\|}
    \end{align}
    Here, $\bm{h}_i$ and $\bm{h}_j$ are the sentence vectors obtained by encoding the sentences $s_i$ and $s_j$, respectively.
    Since the proposed method edits neighbor examples at word level, it is preferable that a neighbor example has common words with a target sentence.
    Therefore, we retrieve $k$ candidate neighbors based on $S_{\mathrm{SentVec}}$, and rerank the $k$ candidates by using the lexical matching with $S_{\mathrm{TFIDF}}(s_i, s_j)$.

%% file: sections/4-exp.tex
\section{Experiment}
    \subsection{Experimental Settings}
    \paragraph{Dataset}
    We used the JRC-Acquis English-German dataset\footnote{\url{https://opus.nlpl.eu/JRC-Acquis.php}}, which is a corpus of legal documents.
    Because the corpus includes quite a few semantically similar sentences, many previous studies utilizing nearest neighbors regard this corpus as the standard benchmark dataset~\cite{li-etal-2016-phrase,gu2018search,xu-etal-2020-boosting,cai-etal-2021-neural}.
    In this experiment, we preprocessed the data after excluding duplicate translation pairs, following \citet{gu2018search}.

    \paragraph{Models}
    We used Transformer~\cite{vaswani2017attention} (base setting) as the strong AR baseline (AR).
    For the NAR baselines, we adopted Levenshtein Transformer~\cite{NEURIPS2019_675f9820} (LevenshteinT), GLAT+CTC~\cite{qian-etal-2021-glancing}, NAT-ITR~\cite{lee-etal-2018-deterministic}, NAT-CRF~\cite{NEURIPS2019_74563ba2}, and CMLM~\cite{ghazvininejad-etal-2019-mask}.
    We used a beam size of four during inference for the AR model.
    Implementing these models using fairseq~\cite{ott-etal-2019-fairseq}\footnote{\url{https://github.com/pytorch/fairseq}}, we trained them with a maximum of 100,000 steps and a total batch size of approximately 32,768 tokens per step.

    \paragraph{Neighbor retrieval}
    We used TFIDF and SentVec+TFIDF ($k$=50) to retrieve neighbors from the datastore and chose the closest sentence for each input sentence.
    The encoder of SentVec+TFIDF is the SBERT model pre-trained on multilingual corpora\footnote{\url{https://huggingface.co/sentence-transformers/distiluse-base-multilingual-cased-v1}}~\cite{reimers-gurevych-2019-sentence}, which performs well in information retrieval tasks.
    We used faiss\footnote{\url{https://github.com/facebookresearch/faiss}}, a fast library for nearest-neighbor retrieval.

    \paragraph{Evaluation}
    We assessed the quality of the generated sentences using a case-sensitive detokenized BLEU using SacreBLEU~\cite{post-2018-call} and ChrF~\cite{popovic-2015-chrf}.
    We also calculated the mean number of iterations and the mean time required to translate a sentence.
    The latter measured the mean latency while the model generated a sequence using a single batch and a single Tesla P100 GPU, similar to \citet{xu-carpuat-2021-editor}.
    Latency includes the time for nearest-neighbor retrieval (for NeighborEdit).
    
    \vspace{3mm}
    A more detailed description can be found in the Appendix.

    \subsection{Experimental Result\label{subsec:main_result}}
    We presented the main results on generation quality and decoding speed in Table~\ref{tab:main_result} and the translation example in  Appendix~\ref{subsec:output_ex}.

    \begin{table*}[t]
        \centering
        \begin{tabular}{cccrrrrp{3mm}}
            \hline\hline
             &\multirow{2}{*}{Model} 
            & \multicolumn{1}{c}{Initial state} & \multirow{2}{*}{BLEU}&\multirow{2}{*}{ChrF}& \multicolumn{1}{c}{Number of} 
            & \multicolumn{2}{c}{ Latency [ms] } \\
            & 
            & \multicolumn{1}{c}{of decoder} & && \multicolumn{1}{c}{iterations}& \multicolumn{2}{c}{\small{(retrieval time)}} \\\hline\hline
            AR
            & Transformer                            
            & \texttt{[BOS]}   & 53.90    & 	72.51     & 30.69        & 300.8         &   \\\hline
          NAR  & GLAT+CTC      
             & $\mathrm{Encoder}(\bm{x})$ &  46.75 &66.53    & 1.00         &    19.8   &         \\\hline
           
            & NAT-ITR            
            & $\bm{x}$ &  38.53   &60.71  & 3.57         &     46.7           \\
            & NAT-CRF           
            & $\texttt{[PAD]},..., \texttt{[PAD]}, \texttt{[EOS]}$ &  26.94   &50.39  & 3.25        &    64.7    &       \\\cline{2-8}
            & \multirow{3}{*}{CMLM}             
            & \multirow{3}{*}{$\texttt{[MASK]}, ..., \texttt{[MASK]}$} &  47.81    &67.55     & 10.00          &    137.8           \\
            &
            &  &  46.98     &67.06   & 6.00          &   88.0             \\
          Iterative &                &            
            & 44.12      &65.17   & 3.00          &    50.6     & \\\cline{2-8}
            NAR& LevenshteinT         
            & \texttt{<s></s>} & 48.36 &69.00&  2.51      & 93.0          &   \\\cline{2-8}
            & \multirow{2}{*}{NeighborEdit} 
            & TFIDF            &  \underline{55.40}&72.59 &  1.66  & 75.4         & \small{(0.17)}\\
            &                                   
            & SentVec+TFIDF          & \underline{\textbf{55.59}} & \textbf{72.74}&  1.65     & 77.0          & \small{(0.08)} \\
            \hline\hline
        \end{tabular}
                \caption{BLEU, the mean number of iterations, and the mean latency [ms] on the JRC-Acquis dataset. \textbf{Boldface} numbers are the top scores among NAR models. The \underline{underline} denotes significant gains over LevenshteinT at $p < 0.05$.
                \label{tab:main_result}}
    \end{table*}
    NeighborEdit outperformed the NAR baselines significantly (more than +7.04 BLEU score and +3.59 ChrF score).
    Furthermore, the mean number of iterations was the lowest among the iterative NAR models.
    NeighborEdit was not the fastest in terms of latency because the proposed method decodes three times (once for deletion and twice for insertion) for one iteration.
    However, it is noteworthy that the proposed method significantly improved the generation quality.
    
    Surprisingly, NeighborEdit outperformed even a strong AR baseline (more than +1.50 BLEU score) on this dataset.
    The mean number of iterations were less than one-eighteenth of the AR baseline, and the latency was reduced by a quarter.
    
    These results indicate that incorporating neighbors into an NAR model is effective; NeighborEdit can consistently improve the generation quality without sacrificing the decoding speed.
    Furthermore, this finding raises another question regarding the way in which nearest neighbors are utilized in the proposed method.
    Therefore, we investigated the effect of the neighbors in the oracle policy and the decoder initialization separately.
    We also explored a \textsc{Rand} baseline that retrieves an example at random (regardless of input) for decoder initialization.

    \begin{table}[]
        \centering
        \begin{tabular}{lr*{1}{S[table-format=2.2]}r}\hline
            Oracle policy   & \multicolumn{3}{c}{Decoder initialization} \\
          for neighbors  &\small{\texttt{<s></s>}}& \multicolumn{1}{c}{$\bm{z}_0$}  &\textsc{Rand}\\\hline
     No ($\pi^*_{\mathrm{t}}$)   & 48.36 &42.05$^\dag$&3.00\\
    Yes ($\pi^*_{\mathrm{t}}+\pi^*_{\mathrm{n}}$) &45.67&55.59 & 44.09\\\hline
        \end{tabular}
                \caption{BLEU scores with or without neighbor in the oracle policy and decoder initialization on JRC-Acquis dataset.
                $\dag$: We displayed the best-performing value using SentVec+TFIDF, although the model can be initialized with TFIDF and SentVec+TFIDF. \label{tab:w_wo_neighbor}}
    \end{table}
    Table~\ref{tab:w_wo_neighbor} shows that the performance significantly improved when we incorporate neighbors from both the oracle policy and decoder initialization.
    Moreover, random decoder initialization severely hurt the performance.
    These results imply that the key to performance improvement is choosing useful neighbors and incorporating them effectively to the oracle policy and decoder initialization.

    \subsection{Experiments on Other Datasets\label{subsec:analysis}}
    
    We also experimented other datasets that have not been well explored in previous studies using nearest neighbors.
    In these experiments, we adopted Levenshtein Transformer as an NAR baseline.

    \subsubsection{Data-to-Text (WikiBio)}
    \label{sec:experiment-wiki-bio}
    We evaluated the effectiveness of NeighborEdit in the task of data-to-text generation on WikiBio dataset\footnote{\url{https://rlebret.github.io/wikipedia-biography-dataset/}}~\cite{lebret-etal-2016-neural}.
    In this task, the model receives a fact table describing a person and generates the biography.
    Because biographies are written in a similar manner, we can assume that certain patterns exist in the target side of this task. 
    To obtain the nearest neighbor, we calculated the similarity score $S_{\text{TM}}$ between two serialized infoboxes $s_i$ and $s_j$, similarly to \citet{wiseman-etal-2021-data}.
    \begin{align}
        S_{\text{TM}}(s_i, s_j) = &F_1(\mathrm{fields}(s_i), \mathrm{fields}(s_j)) \label{eq:stm} \\ \notag
        &+0.1F_1(\mathrm{values}(s_i), \mathrm{values}(s_j)) 
    \end{align}
    Here, $\mathrm{fields}(s)$ extracts the field types (e.g., `name') from the infobox $s$, $\mathrm{values}(s)$ extracts the unigrams that appear as values in the infobox $s$ (e.g., `Obama'), and $F_1$ presents an F1-score.
    We call this retrieval TableMatch.
    We trained the models with a maximum of 300,000 steps and a batch size of approximately 16,384 tokens per step.
    We evaluated the models using BLEU, NIST\footnote{We used this evaluation script: \url{https://github.com/tuetschek/e2e-metrics}.} and the mean number of iterations.
    
    \begin{table}[t]
        \centering
        \begin{tabular}{lrrr}
            \hline
          Model &BLEU&NIST&Iteration\\\hline
          Transformer  &46.91&9.48&23.70\\\hline
          LevenshteinT &43.09&8.79&2.25\\
          NeighborEdit&\textbf{\underline{44.96}}&\textbf{9.24}&\textbf{2.08}\\\hline
        \end{tabular}
                \caption{Experimental results on the WikiBio dataset.  \textbf{Boldface} numbers are the top scores among NAR models.  The \underline{underlined} values denote significant gains over LevenshteinT at $p < 0.05$.\label{tab:WikiBio}}
    \end{table}
    Table~\ref{tab:WikiBio} reports the evaluation result.
    NeighborEdit improved the BLEU score compared to the NAR baseline (+1.87 points) and narrowed the performance gap with the AR baseline with fewer iterations.
    This demonstrates that the proposed method was effective on this dataset.

    \subsubsection{Machine Translation (WMT'14)}
    Thus far, we examined the performance of NeighborEdit on the datasets where nearest neighbors may be effective.
    In this section, we investigate the performance on the popular WMT'14 English-German dataset~\cite{bojar-etal-2014-findings}.
    Unlike JRC-Acquis, this dataset covers a wide range of topics.
    We downloaded the dataset and preprocessed it using the fairseq scripts\footnote{\url{https://github.com/pytorch/fairseq/tree/main/examples/translation\#WMT'14-english-to-german-convolutional}}.
    We trained the models with a maximum of 300,000 steps and a batch size of approximately 65,536 tokens per step.

    \begin{table}[t]
        \centering
        \begin{tabular}{lrrr}
            \hline
          \multicolumn{2}{c}{Model}  &BLEU&Itr.\\\hline
          Transformer  &&27.06 & 28.10\\
          +\textsc{Concat} &&26.91& 28.02\\\hline
          LevenshteinT &&23.98	&	2.83\\ 
          \multirow{2}{*}{NeighborEdit}&\small{TFIDF}&24.17	&2.60 \\
          &\small{SentVec+TFIDF} &\underline{\textbf{24.61}}	&	3.05\\\hline
        \end{tabular}
                \caption{BLEU scores and mean number of iterations on the WMT'14 En-De dataset.  \textbf{Boldface} numbers are the top scores among NAR models. The \underline{underlined} values denote significant gains over LevenshteinT at $p < 0.05$.\label{tab:wmt_main_result}}
    \end{table}
    Table~\ref{tab:wmt_main_result} demonstrates that the proposed method using SentVec+TFIDF outperforms Levenshtein Transformer (+0.63 points).
    However, the performance gain of BLEU scores was smaller than that in the other datasets, and the number of iterations was larger.
    Although we also explored an AR model where a source sentence and its neighbor are concatenated in the Transformer architecture (+\textsc{Concat}), we could not observe a performance improvement.
    Therefore, we suspect that the WMT'14 En-De dataset may have a different characteristic to JRC-Acquis and WikiBio.
    
    \subsection{Analysis on Neighbor Examples\label{subsubsec:analysis}}
\begin{figure*}[hbt!]
            \begin{subfigure}{.475\linewidth}
                \includegraphics[width=\linewidth]{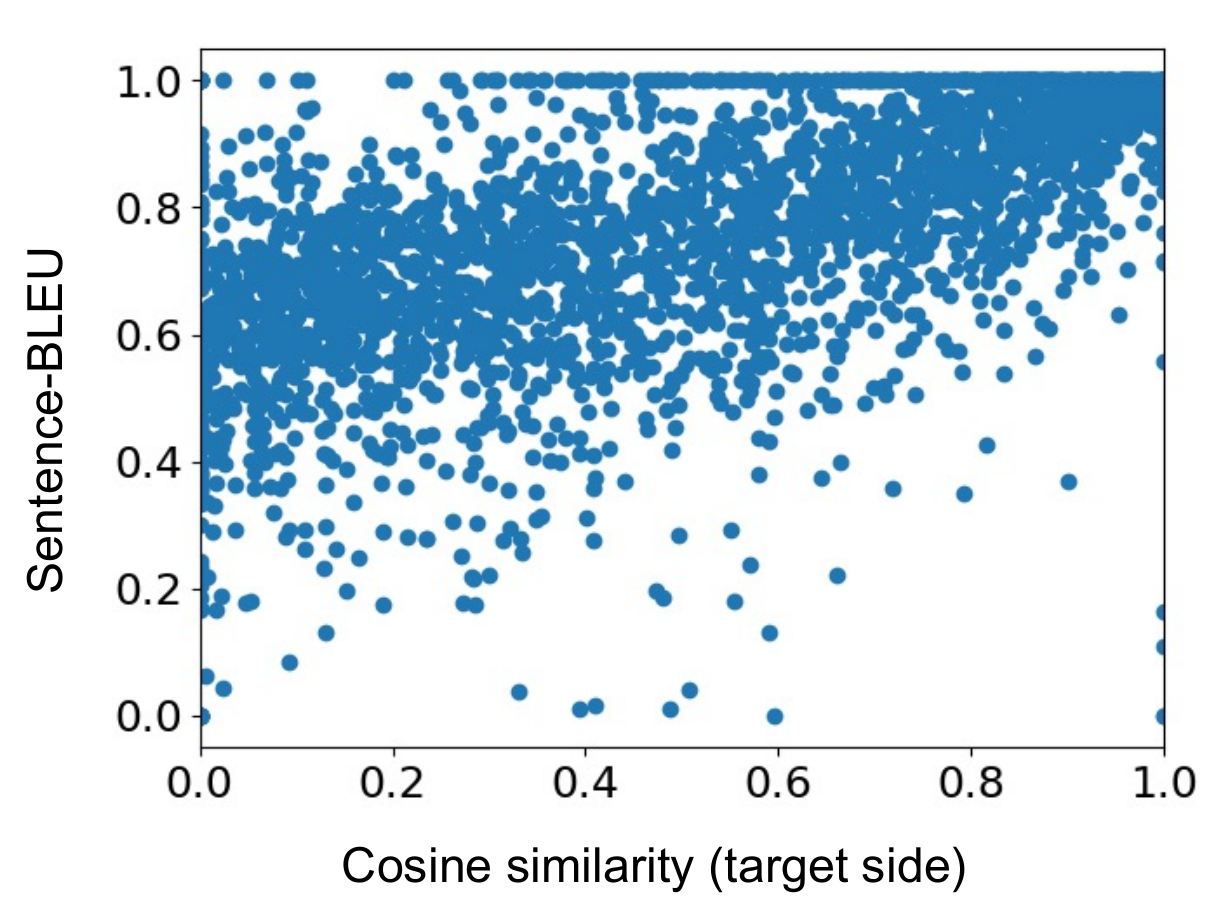}
                \caption{JRC-Acquis En-De\label{fig:scatter_jrc}}
            \end{subfigure}\hfill %
            \begin{subfigure}{.475\linewidth}
                \includegraphics[width=\linewidth]{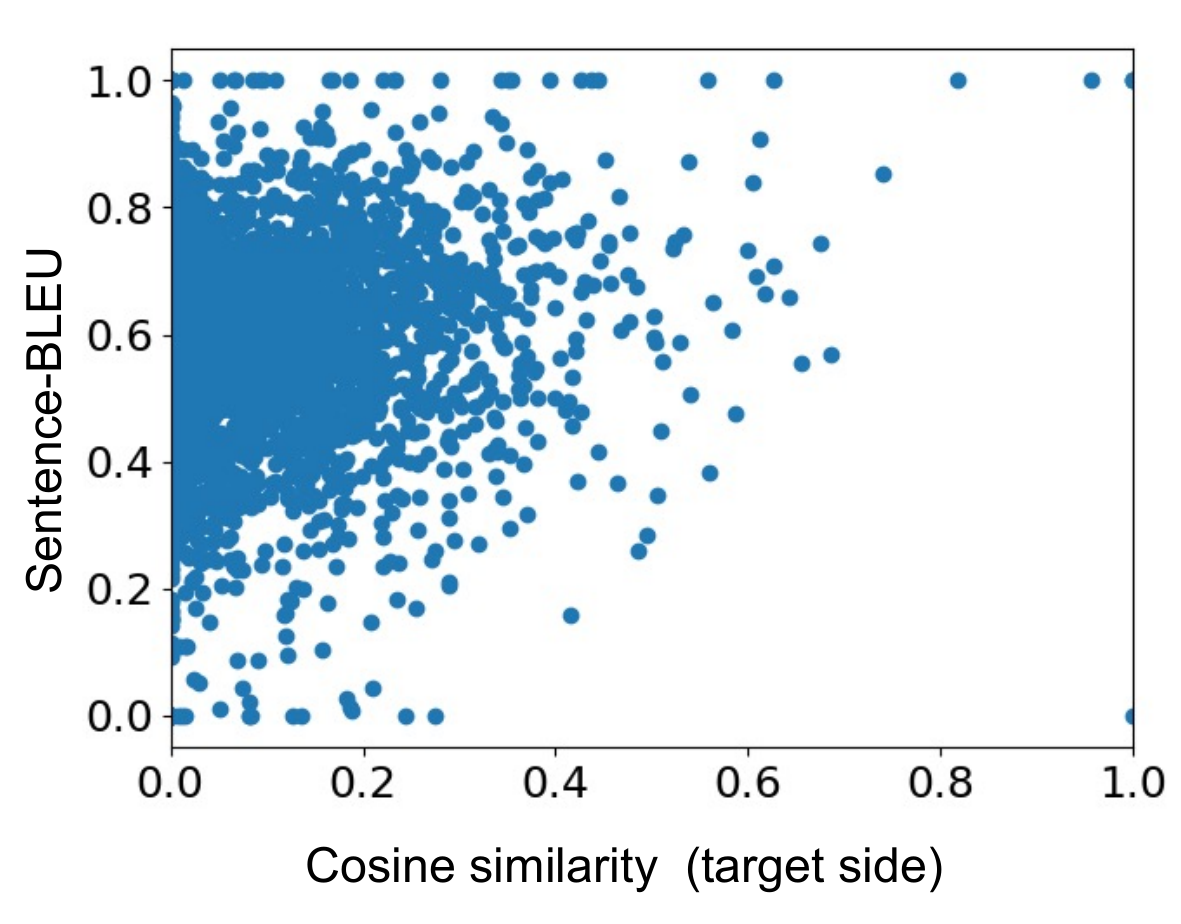}
                \caption{WMT'14 En-De\label{fig:scatter_wmt}}
            \end{subfigure}
            \medskip %
            \begin{subfigure}{.475\linewidth}
                \includegraphics[width=\linewidth]{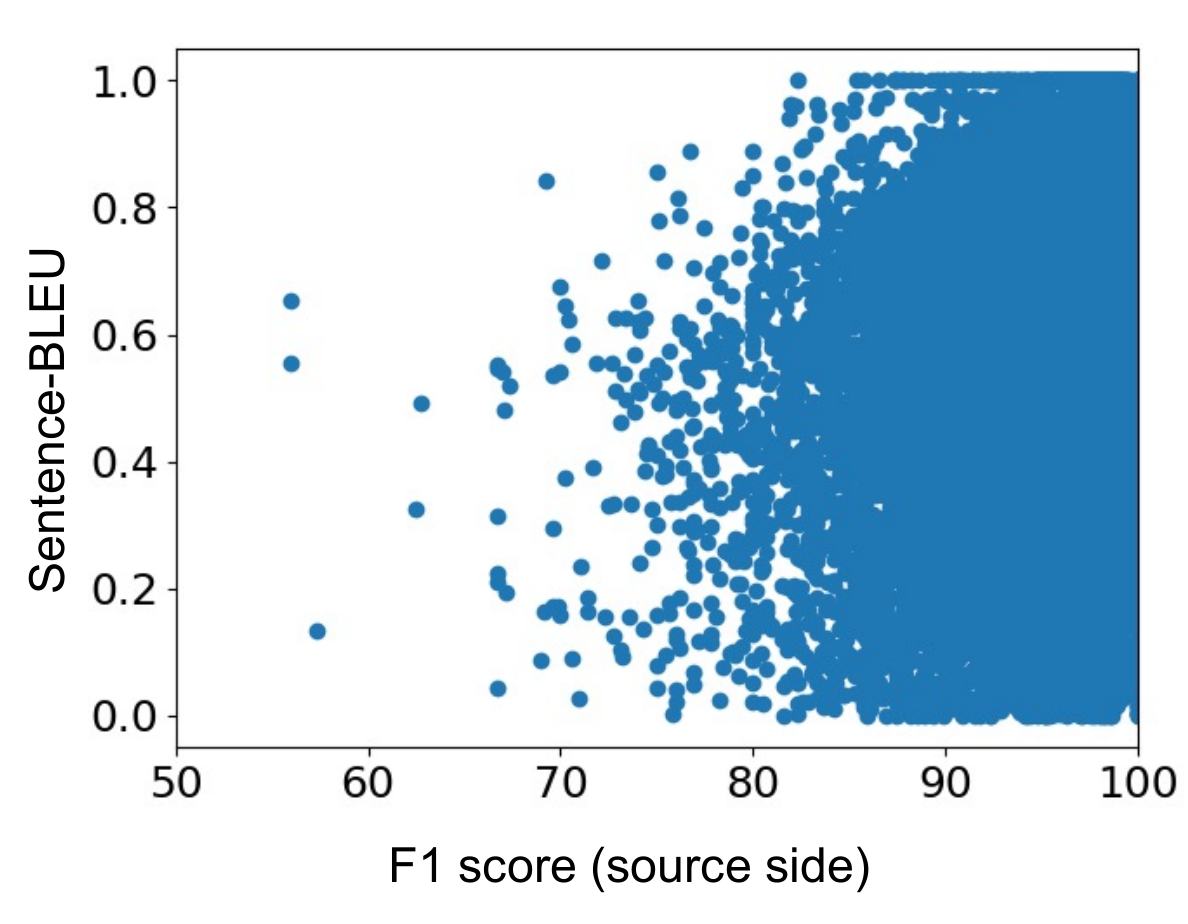}
                \caption{WikiBio (source side) \label{fig:scatter_wikibio_source}}
            \end{subfigure}\hfill %
            \begin{subfigure}{.475\linewidth}
                \includegraphics[width=\linewidth]{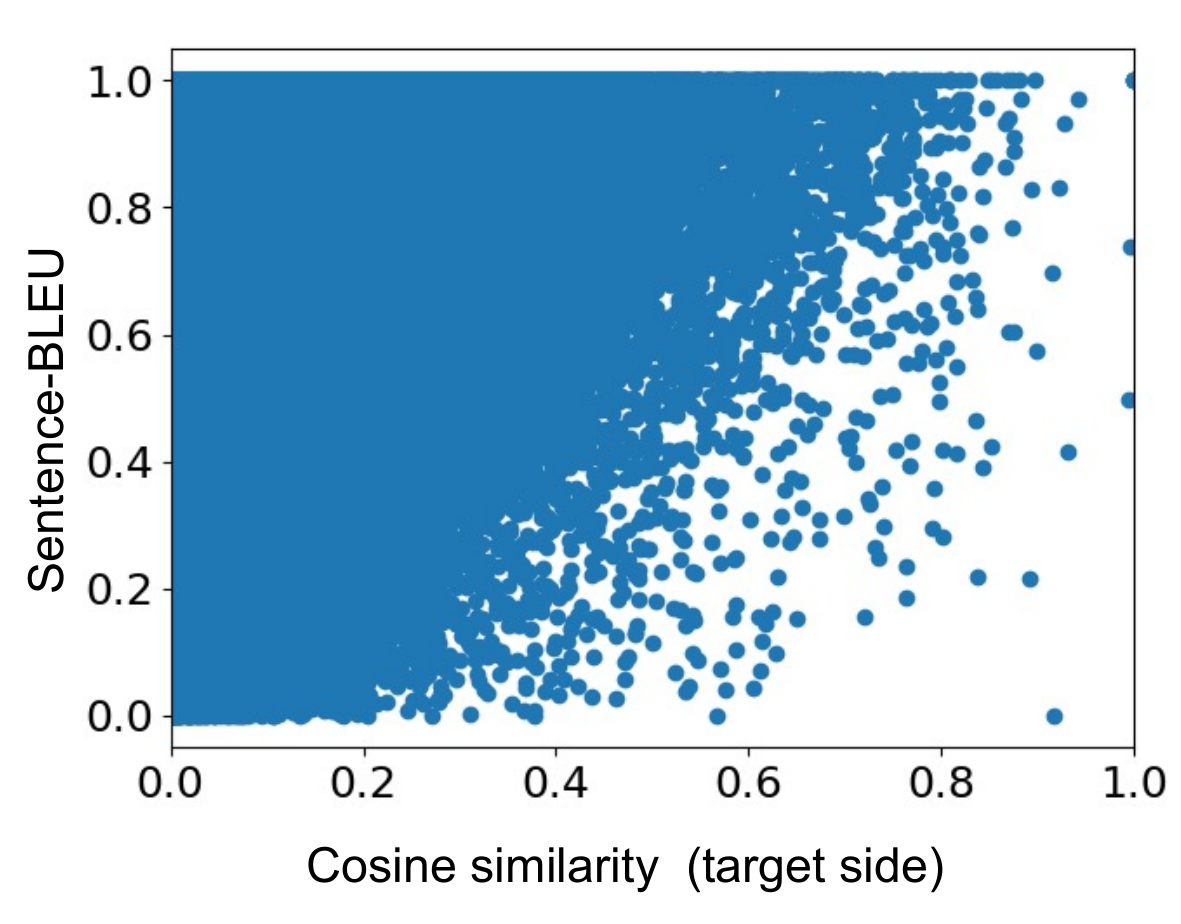}
                \caption{WikiBio (target side)\label{fig:scatter_wikibio_target}}
            \end{subfigure}
            \caption{Scatter plot depicting the similarity between the example and its neighbor ($x$-axis) and sentence-BLEU  ($y$-axis) in each development set.\label{fig:scatter}}
        \end{figure*}

    Figure~\ref{fig:scatter} displays scatter plots between the similarity of an example and its neighbor ($x$-axis) and the sentence-BLEU of its translation ($y$-axis). 
    For JRC-Acquis and WMT'14 datasets, the $x$-axis presents the cosine similarity of tf-idf vectors between a target sentence and its neighbor.
    However, the similarity measures of the source and target sides of the WikiBio dataset are completely different; therefore, the $x$-axis of Figure~\ref{fig:scatter_wikibio_source} shows F1 scores of overlapping field names and values (Equation~\ref{eq:stm}) (source side); and Figure~\ref{fig:scatter_wikibio_target} shows the cosine similarity of tf-idf vectors between a target sentence and its neighbor (target side).

    Figure~\ref{fig:scatter_jrc} shows that the JRC-Acquis dataset contains many similar examples; the mean of cosine similarity is 0.521. 
    We observe the tendency that BLEU scores are larger when retrieved examples have higher similarity values to target sentences.
     
    In contrast, Figure~\ref{fig:scatter_wmt} clearly indicates that the WMT'14  dataset does not include similar examples; the mean of cosine similarity is 0.105.
    This implies that NeighborEdit must edit retrieved examples a lot, beginning with distant examples.
    This difficulty has already been observed in the larger number of iterations in Table \ref{tab:wmt_main_result}.
    In other words, NeighborEdit was not so effective for WMT'14 because most examples were dissimilar and few examples with a high similarity were useful.
    
    In the WikiBio dataset, the mean similarity on the source side is as high as 99.15 (Figure~\ref{fig:scatter_wikibio_source}) because two infobox tables can be similar only if these tables include the same field names.
    In contrast, the mean similarity on the target side is 0.130 (Figure~\ref{fig:scatter_wikibio_target}).
    These plots indicate that the similarity distributions on the source and target sides are quite different because different similarity metrics are used for infobox tables and sentences.
    We followed the previous study to design the similarity metric of the source side in Section \ref{sec:experiment-wiki-bio}.
    However, it may be necessary to reconsider the similarity metric for the source side so that it reflects similar patterns of biographies, for example, by focusing more on the occupations of entities.

%% file: sections/5-related_works.tex
    \section{Related Work}

    \subsection{Non-autoregressive Generation}
    Text generation using NAR decoding has attracted attention in recent years~\cite{ghazvininejad-etal-2019-mask,lee-etal-2020-iterative, NEURIPS2019_675f9820, stern2019insertion, lee-etal-2018-deterministic}.
    Common approaches for improving the performance are iterative decoding and knowledge distillation using AR models.
    A number of researchers have proposed non-left-to-right decoding, utilizing parallelism in NAR models~\cite{ghazvininejad-etal-2019-mask,NEURIPS2019_675f9820,stern2019insertion}.
    Inspired by these approaches, we incorporated the nearest neighbor into the iterative decoding process to guide the generation.
    
    \subsection{Text Generation using Neighbors }

    \citet{gu2018search} were the first to use nearest neighbors in an attention-based encoder-decoder model. Other studies utilized the neighbors at the token-level~\cite{khandelwal2021nearest, zheng-etal-2021-adaptive}, chunk-level~\cite {borgeaud2021improving}, and sentence-level~\cite{peng-etal-2019-text, cao-etal-2018-retrieve} for text generation.
    Our approach is motivated primarily by these successes.
    While token-level and chunk-level retrieval can obtain less noisy (i.e., close) neighbors, they need to repeat a retrieval for each time step, which slows down training and inference.
    In contrast, our method does not suffer from the slowdown, retrieving neighbors only once per input.
    We adjusted how neighbors were utilized during training and inference to reduce the influence of noisy neighbors retrieved at sentence level.
    
    Some previous studies concatenated a source sentence and its neighbors as input or used two encoders for them~\cite{xu-etal-2020-boosting, bulte-tezcan-2019-neural,borgeaud2021improving}.
    Although these methods are effective, they require additional mechanisms and parameters.
    In contrast, we incorporated neighbors without introducing additional components or parameters but only by changing the training strategy.
    To the best of our knowledge, this is the first study to incorporate neighbors into an NAR model.

%% file: sections/6-conclusion.tex
\section{Conclusion}
This paper proposes NeighborEdit, which utilizes the nearest neighbors in an NAR decoder based on edit operations.
The experimental results showed that NeighborEdit could improve the quality of the generated sentences with fewer iterations than existing NAR models on all datasets.
We expect that NeighborEdit will be beneficial in domains with writing patterns, such as patents and parliamentary proceedings.
Future work will include the design of improved oracle policies that can reproduce intermediate sequences during inference as well as the development of improved algorithms for sentence-level neighbor retrieval.

\section*{Acknowledgments}
This work was supported by JSPS KAKENHI Grant Number 21J13602.
These research results were obtained from the commissioned research (No. 225) by National Institute of Information and Communications Technology (NICT), Japan.